\documentclass[journal]{IEEEtran}
\usepackage{amsfonts}
\usepackage{amsmath}
\usepackage{balance}
\usepackage{graphicx}
\usepackage{multirow}
\usepackage{subfigure}
\usepackage{xcolor}

\renewcommand\fbox{\fcolorbox{gray}{white}}
\begin{document}
\bstctlcite{IEEEexample:BSTcontrol}

\title{Nutrition Estimation for Dietary Management: A Transformer Approach with Depth Sensing}

\author{Zhengyi Kwan,~\IEEEmembership{Student Member,~IEEE,}~Wei Zhang,~\IEEEmembership{Member,~IEEE,}~Zhengkui Wang~\IEEEmembership{Member,~IEEE,}\\Aik Beng Ng~and Simon See~\IEEEmembership{Senior Member,~IEEE}
\thanks{Manuscript received January 1, 2000; revised January 1, 2000. This work was supported in part by A*STAR under its MTC Programmatic (Award M23L9b0052), MTC Individual Research Grants (IRG) (Award M23M6c0113), the Ministry of Education, Singapore, under the Academic Research Tier 1 Grant (Grant ID: GMS 693), SIT’s Ignition Grant (STEM) (Grant ID: IG (S) 2/2023 – 792), and Singapore’s Economic Development Board - Industrial Postgraduate Programme, in conjunction with NVIDIA AI Technology Center and SIT. (\textit{Corresponding author: Wei Zhang})}
\thanks{Zhengyi Kwan is with both the NVIDIA AI Technology Center, Singapore 038988, and the Information and Communications Technology Cluster, Singapore Institute of Technology, Singapore 138683 (e-mail: kwanz@nvidia.com and 1800744@sit.singaporetech.edu.sg).}
\thanks{Wei Zhang and Zhengkui Wang are with the Information and Communications Technology Cluster, Singapore Institute of Technology, Singapore 138683 (e-mail: \{wei.zhang, zhengkui.wang\}@singaporetech.edu.sg).}
\thanks{Aik Beng Ng and Simon See are with NVIDIA AI Technology Center, Singapore 038988 (e-mail: \{aikbengn, ssee\}@nvidia.com).}
}

% The paper headers
\markboth{IEEE Transactions on Multimedia, Vol. xx, No. xx, January 2024}%
{Shell \MakeLowercase{\textit{et al.}}: Bare Demo of IEEEtran.cls for IEEE Journals}

\maketitle

\begin{abstract}
Nutrition estimation is crucial for effective dietary management and overall health and well-being. Existing methods often struggle with sub-optimal accuracy and can be time-consuming. In this paper, we propose NuNet, a transformer-based \underline{net}work designed for \underline{nu}trition estimation that utilizes both RGB and depth information from food images. We have designed and implemented a multi-scale encoder and decoder, along with two types of feature fusion modules, specialized for estimating five nutritional factors. These modules effectively balance the efficiency and effectiveness of feature extraction with flexible usage of our customized attention mechanisms and fusion strategies. Our experimental study shows that NuNet outperforms its variants and existing solutions significantly for nutrition estimation. It achieves an error rate of 15.65\%, the lowest known to us, largely due to our multi-scale architecture and fusion modules. This research holds practical values for dietary management with huge potential for transnational research and deployment and could inspire other applications involving multiple data types with varying degrees of importance.
\end{abstract}


\IEEEpeerreviewmaketitle

\section{Introduction}
% context: dietary and nutrition estimation
% problem: how to estimate nutrition
\IEEEPARstart{D}{ietary} management is important for maintaining overall health and well-being. It involves making choices regarding food selection and portions to ensure our bodies receive necessary nutrients without excess. A balanced diet, rich in essential nutrients such as carbohydrates and protein, is vital for supporting optimal bodily functions and promoting health. Conversely, an imbalanced diet or excessive food intake can lead to a range of harmful effects such as obesity, diabetes, chronic diseases, and even mental health issues \cite{tahir2021comprehensive}. Research has demonstrated the effectiveness of dietary management in reducing the risks of various health issues. For example, combined with increased physical activities, dietary management can reduce the likelihood of developing diabetes significantly by nearly 60\% \cite{knower2004diabetes}. A key aspect of dietary management is nutrition estimation. It serves as a valuable tool or method for assessing the nutritional contents of foods, enabling caloric intake management, and identifying potential nutrient deficiencies. 

% existing effort/gap: chart paper; ai
Nutrition estimation is a well-established research field with several traditional approaches available. These include food comparison tables, which provide estimates of the nutritional content of various foods. A similar concept is the nutrition database, which offers better flexibility and accessibility. Measuring cups and spoons are also commonly used tools for nutrition estimation. However, these methods have inherent limitations. They are not accurate, especially for non-trained individuals, as food benchmarks are often too generalized with limited specificity. Besides, estimations can be subjective, due to reasons like memory biases and inconsistent visual interpretations. Moreover, the process may not be user-friendly and time-consuming, especially when frequent and regular estimations are required.

The advancement in smartphones and deep learning over the past decade has offered a new angle for food image analysis \cite{liu2024canteen, li2024multi}. With the capability of capturing food images, smartphones generate food images that are typically RGB-based with the texture and color of the scene. Various machine learning (ML) models can then be used to analyze food images and correlate their visual features with specific nutrients. Once trained, these models can be used for nutrition estimation. To facilitate training, nutrition datasets \cite{thames2021nutrition5k} have been collected and open-sourced. Among the ML models, convolution neural networks (CNNs) like InceptionV2 \cite{szegedy2016rethinking,lo2020image} were studied first, and recent works also applied transformer-based models \cite{shao2022rapid} for achieving improved estimation accuracy. Compared to traditional approaches, these image-based solutions are generally more objective and user-friendly (e.g., non-intrusive and rapid). However, they face challenges also such as unsatisfactory estimation accuracy and application limitations like occlusion. 

% idea: transformer and depth
In this paper, we propose to enhance nutrition estimation by complementing RGB with depth information and analyzing both with the latest transformer architecture. Depth information offers essential spatial and structural insights, beneficial for accurate and robust image analysis in complex scenes like food environments. With nowadays consumer electronics, depth information is often easily accessible. Many smartphones support depth APIs which utilize sensors such as time-of-flight, Lidar, and stereo vision to create depth maps. For phone models lacking such APIs, the depth maps can still be generated with alternative methods like deep learning monocular depth estimation. While researchers have noticed the potential of incorporating depth into nutrition estimation, the information utilization remains elementary \cite{thames2021nutrition5k,shao2023vision, vinod2022image}. Effectively utilizing both RGB and depth data requires advanced ML models and the transformer can be served as a core architecture. In many applications, transformer-based models \cite{dosovitskiy2020image} often outperform CNN-based models with their ability to understand the global context and dynamic focus on the most relevant information. Recent transformers have been explored for nutrition estimation \cite{shao2022rapid}, but none have been designed to be depth-aware and the performance is sub-optimal. 

% method
We propose NuNet, a transformer \underline{net}work specialized for \underline{nu}trition estimation that utilizes both RGB and depth data. The NuNet architecture consists of three building blocks, an encoder, a feature fusion module, and a decoder. To process RGB and depth information simultaneously, we design a multi-scale encoder with two sequences of transformer blocks. Each scale has two parallel blocks for RGB and depth data. These blocks interact with each other, facilitated by our feature fusion modules. We design two types of fusion modules. One is lightweight for integrating features at each encoder scale, and the other has more complex internal mechanisms for enhanced feature fusion at the final encoder scale. Finally, a multi-scale decoder is configured to align with the encoder and feature fusion scales, and produce the final nutrition estimation. In summary, we have the following main contributions in this paper.
\begin{itemize}
    \item We proposed NuNet, a multi-scale transformer architecture specialized for nutrition estimation, which incorporates both RGB and depth data. 
    \item We designed two feature fusion modules to integrate RGB and depth features at each scale, and integrated them into the multi-scale NuNet.
    \item We conducted experimental study and showcased NuNet's performance of nutrition estimation, achieving a 15.65\% error rate, the lowest to our knowledge. 
\end{itemize} 

The impact of NuNet is manifold. NuNet demonstrates the effectiveness of integrating depth information and transformer architecture into nutrition estimation. Its remarkable performance contributes to improved dietary management for health and well-being. It also sheds light on diverse applications such as smart manufacturing and building management.

The rest of the paper is organized as follows. The system architecture of NuNet is presented in Section \ref{sec:system}. Section \ref{sec:method} describes our methodology in detail and Section \ref{sec:exp} presents an experimental study of NuNet as well as result discussions. Finally, we conclude the paper in Section \ref{sec:conclusion}.

\section{System Architecture}
\label{sec:system}
In this section, we introduce the system architecture of NuNet for nutrition estimation. 

\begin{figure}
\includegraphics[width=0.48\textwidth, trim={0 30em 43em 0}, clip]{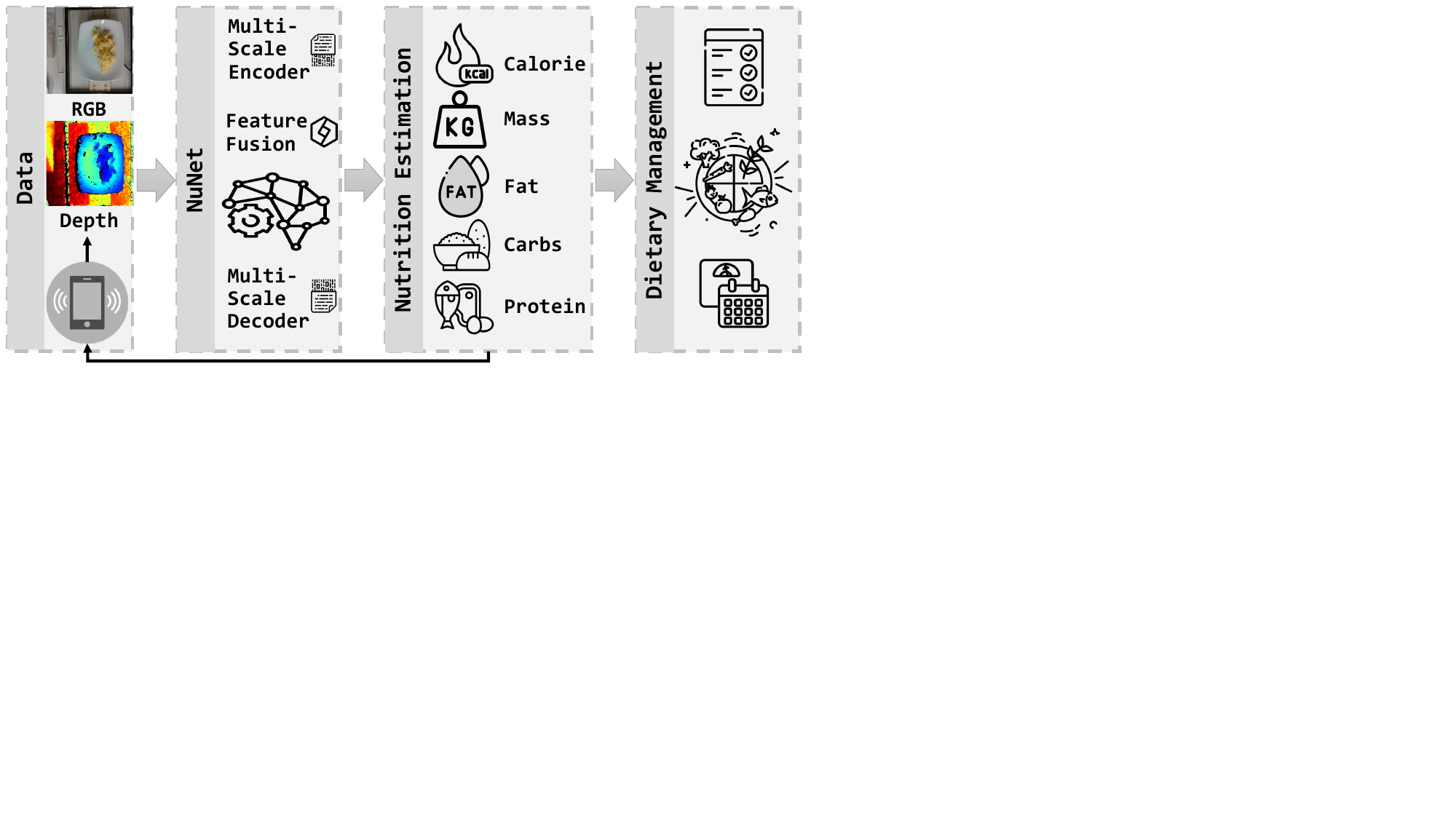}
\caption{An illustration of the system architecture of NuNet. A smartphone with depth sensing captures both RGB and depth images, which are processed by our NuNet for nutrition estimation. The estimation of key nutritional factors is shared with the users and utilized for enhanced dietary management.}
\label{fig:system}
\end{figure}

\subsection{Overview of ML-based Nutrition Estimation}
Fig. \ref{fig:system} shows an illustration of our NuNet system. Typically, a user captures a photo of a meal using a smartphone or another camera-equipped device. The food photos are then processed by ML-based models to estimate nutritional content (e.g., food nutrition and portion sizes). The results of the estimation are subsequently saved automatically in a data repository and shared with the user for smart dietary management. This section first describes the input and output of NuNet and the methodology of processing the input and deriving the output will be presented in Section \ref{sec:method}.

\begin{figure}
\centering
\def \heightData{0.8in}
\def \widthData{1.1in}
\subfigure[Sample 1 - RGB]{\fbox{\includegraphics[height=\heightData, width=\widthData]{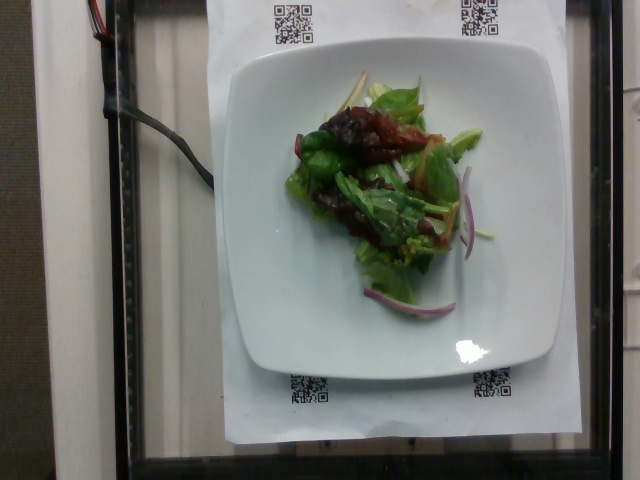}}\label{fig:dish1RGB}}
\hspace{-0.25em}
\subfigure[Sample 2 - RGB]{\fbox{\includegraphics[height=\heightData, width=\widthData]{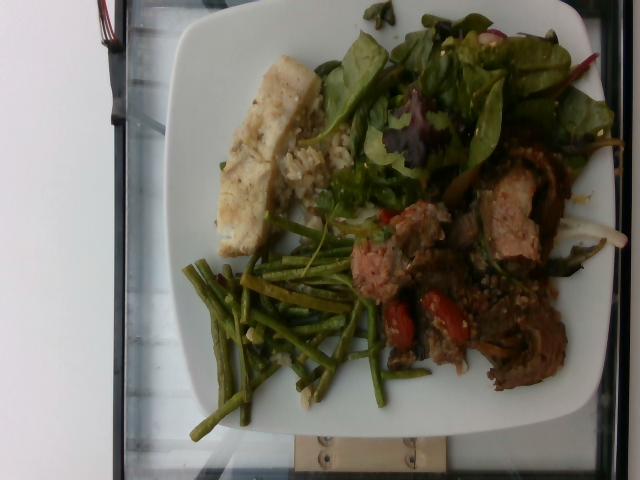}}\label{fig:dish2RGB}}
\hspace{-0.25em}
\subfigure[Sample 3 - RGB]{\fbox{\includegraphics[height=\heightData, width=\widthData]{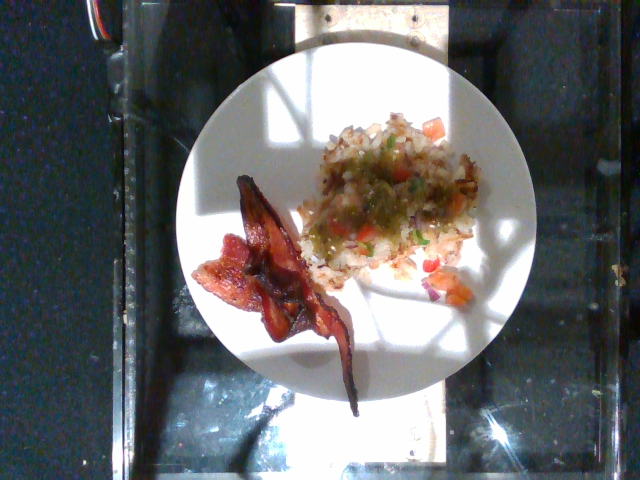}}\label{fig:dish3RGB}}
\subfigure[Sample 1 - Depth]{\fbox{\includegraphics[height=\heightData, width=\widthData]{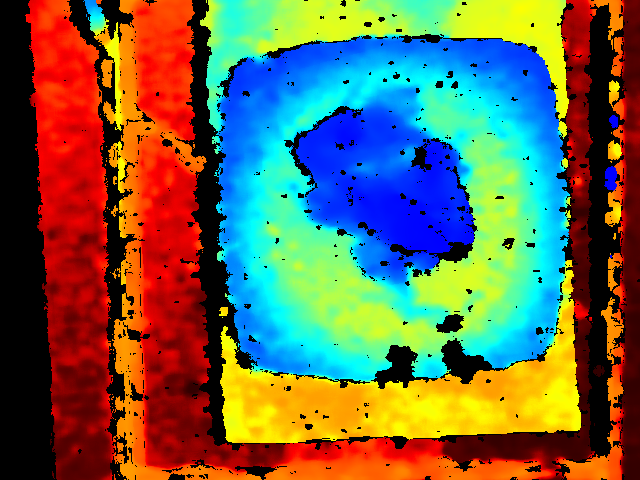}}}\label{fig:dish1D}
\hspace{-0.25em}
\subfigure[Sample 2 - Depth]{\fbox{\includegraphics[height=\heightData, width=\widthData]{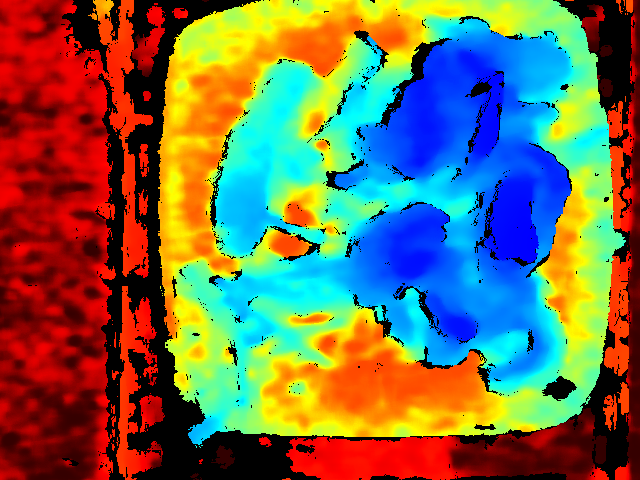}}}\label{fig:dish2D}
\hspace{-0.25em}
\subfigure[Sample 3 - Depth]{\fbox{\includegraphics[height=\heightData, width=\widthData]{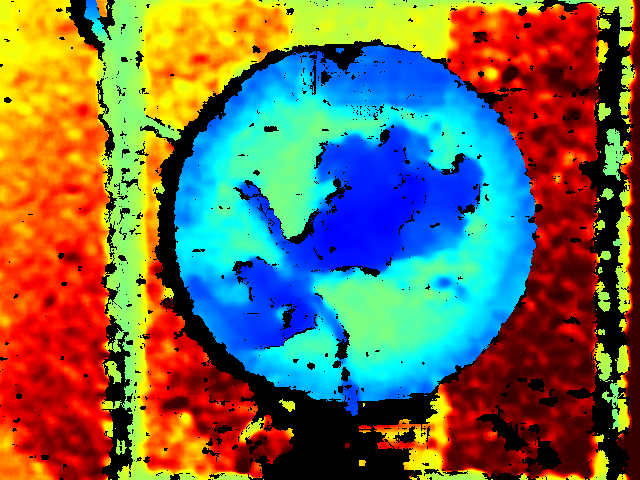}}}\label{fig:dish3D}
\caption{Sample food images with both RGB and depth information from the \texttt{Nutrition5k} dataset \cite{thames2021nutrition5k}.}
\label{fig:sample-images}
\end{figure}

\subsection{Input with RGB and Depth}
For input, cameras typically provide colored images. Some users may not be aware of the recent advancements in smartphone technologies, which often include depth-sensing capabilities. These cameras can simultaneously capture both RGB and depth images in a single shot. A few sample RGB and depth food images are shown in Fig. \ref{fig:sample-images}. In this research, we plan to utilize both. We define each input as a tuple $(\mathbf{x}, \mathbf{d})$, representing a single food capture, where $\mathbf{x}$ and $\mathbf{d}$ correspond to the colored and depth images, respectively. The colored image $\mathbf{x}$ has three channels for RGB and the depth image $\mathbf{d}$ contains one channel only for depth. To standardize the process, both images are resized to a width $W$ and a height $H$, and we have $\mathbf{x} \in \mathbb{R}^{W \times H \times 3}$ and $\mathbf{d}\in \mathbb{R}^{W \times H}$.

\begin{figure*}
\centering
\includegraphics[width=0.98\textwidth, trim={0 16em 0 0}, clip]{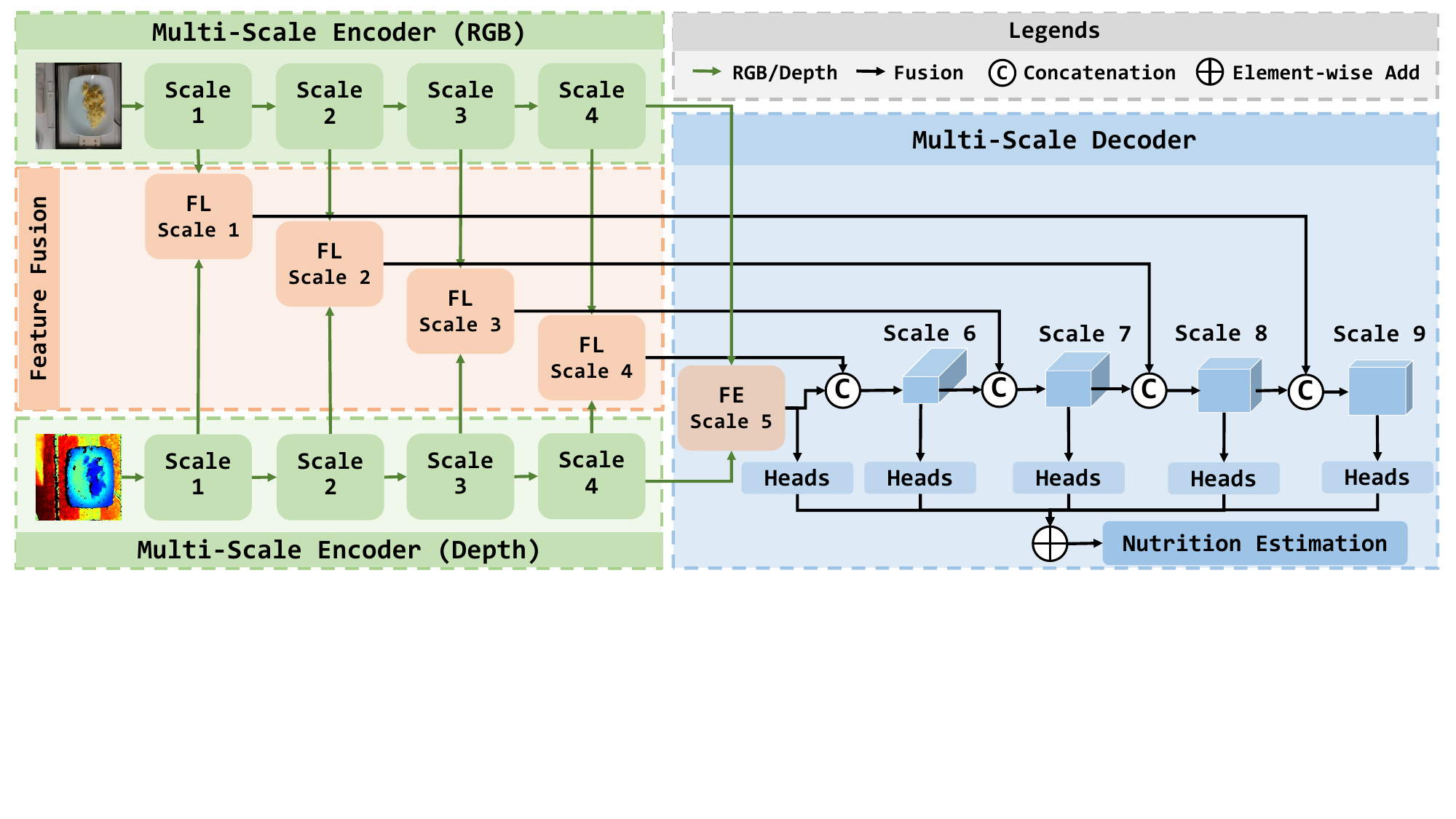}
\caption{An illustration of the system architecture of NuNet. NuNet consists of three key components, including a multi-scale encoder, a feature fusion module, and a multi-scale decoder. It utilizes both RGB and depth images as input and analyzes the data using its transformer architecture. Finally, NuNet generates the nutrition estimation values for dietary management.}
\label{fig:model}
\end{figure*}

\subsection{Output for Nutrition Estimation}
The output of NuNet is nutrition estimation, which is crucial for dietary management. We consider several key nutritional factors, including calories, mass, fat, carbohydrates (or carbs), and protein. For an input tuple $(\mathbf{x}, \mathbf{d})$, let $\mathbf{y}$ represent the nutritional value of the food, where $\mathbf{y} = \{y^j\}_{j=1,\ldots,k}$ correspond to the $k$ aforementioned factors. These factors are represented in order as $y^\text{cal}$, $y^\text{mass}$, $y^\text{fat}$, $y^\text{carb}$, $y^\text{protein}$ for calorie, mass, fat, carbs, and protein values, respectively. Here, $\mathbf{y}$ is considered as the ground-truth for each nutrition task, and $\hat{\mathbf{y}}$ is the predicted value for the same task. The key objective of this research is to develop an ML model that can accurately estimate the nutritional values based on the input images. We present the technical details of NuNet below.

\section{Methodology and Model Description}
\label{sec:method}
We present our methodology and NuNet models in this section. We start with an overview of the model, followed by a detailed description of the model's key components. An illustration of NuNet is shown in Fig. \ref{fig:model}.

\subsection{NuNet Overview}
NuNet, which converts food image input to nutrition estimation output, consists of three key components, an encoder, a feature fusion module, and a decoder. The encoder is important for contextual understanding of the input and feature extraction, and we design it based on the Swin Transformer (SwinT) Base  \cite{liu2021swin}. A challenge of NuNet is the dual-input with both RGB and depth images, where the strategies of utilizing such dual-input can impact the overall performance significantly. Thus, developing an effective strategy for fusing RGB and depth features is a core research task in this paper. Specifically, we have one transformer to process RGB images and it is decomposed into four scales $s_1,\ldots,s_4$. In parallel, we have another transformer for depth images and we align it with the RGB transformer with the same scales $s_1,\ldots,s_4$. We collect the output of each scale for both RGB and depth images, denoted as features $\mathbf{f}_i^\text{R}$ and $\mathbf{f}_i^\text{D}$, respectively, for scale $s_i$ where $1\leq i \leq 4$. 

Our strategy is to adopt the multi-scale feature fusion to utilize the features mentioned earlier. We apply a standard fusion process for the four scales to process $\mathbf{f}_i^\text{R}$ and $\mathbf{f}_i^\text{D}$ and produce $\mathbf{f}_i^\text{F}$ for scale $s_i$. We follow the common assumption that the scales closer to the output are more specialised and correlated to application, e.g., nutrition estimation in our case. We give special attention to the last scale $s_4$, for which we design an enhanced fusion module and generate feature $\mathbf{f}_5^\text{F}$, which is tagged as scale 5 for presentation clarity.

The fusion features are processed by the decoder to generate the final nutrition estimation. The decoder operates at multiple scales to match the five scales of feature fusion. We first derive the estimates from the enhanced feature $\mathbf{f}_5^\text{F}$ which passes through linear heads for the five nutrients. We deploy four scales in the decoder from $s_6$ to $s_{9}$, where scale $s_i$ for $6\leq i\leq 9$ incorporates the output feature of scale $s_{i-1}$ and concatenates it with the feature from scale $s_{10-i}$. Linear heads are then appended to each of these scales to calibrate the nutrition estimation. Finally, the final nutrition estimation, as the aggregated information from all decoder scales, serves as the output of the whole model.

\subsection{Multi-Scale Encoder}
SwinT \cite{liu2021swin} is developed on top of the popular Vision Transformer (ViT) \cite{dosovitskiy2020image} and has been demonstrated effective in image processing tasks. We use it as our encoder subject to necessary customization. Specifically, we run two parallel SwinTs, one processes RGB images in four scales and the other handles depth images with the same settings. We follow SwinT's implementation of non-overlapping windows, which improves the computational complexity of the self-attention mechanism. The implementation brings two multi-head self-attention (MSA) mechanisms including the window-based W-MSA followed by shift-window-based SW-MSA. Both mechanisms capture the spatial relationship between patches within the window instead of the whole image and the global attention is achieved by shifting the windows based on the optimized window partition layouts.

\subsection{Feature Fusion}
NuNet processes both RGB and depth images in parallel using its SwinT-based encoder and strategically utilizing the features derived from these images is critical. In this part, we present our feature fusion strategies, which balance fusion effectiveness and computational efficiency. Specifically, we develop two types of fusion modules, including lightweight and enhanced fusion denoted as FL and FE, respectively. We will detail the technical aspects of both modules as follows. 

\begin{figure}
\centering
\includegraphics[width=0.48\textwidth, trim={0 41em 54em 0}, clip]{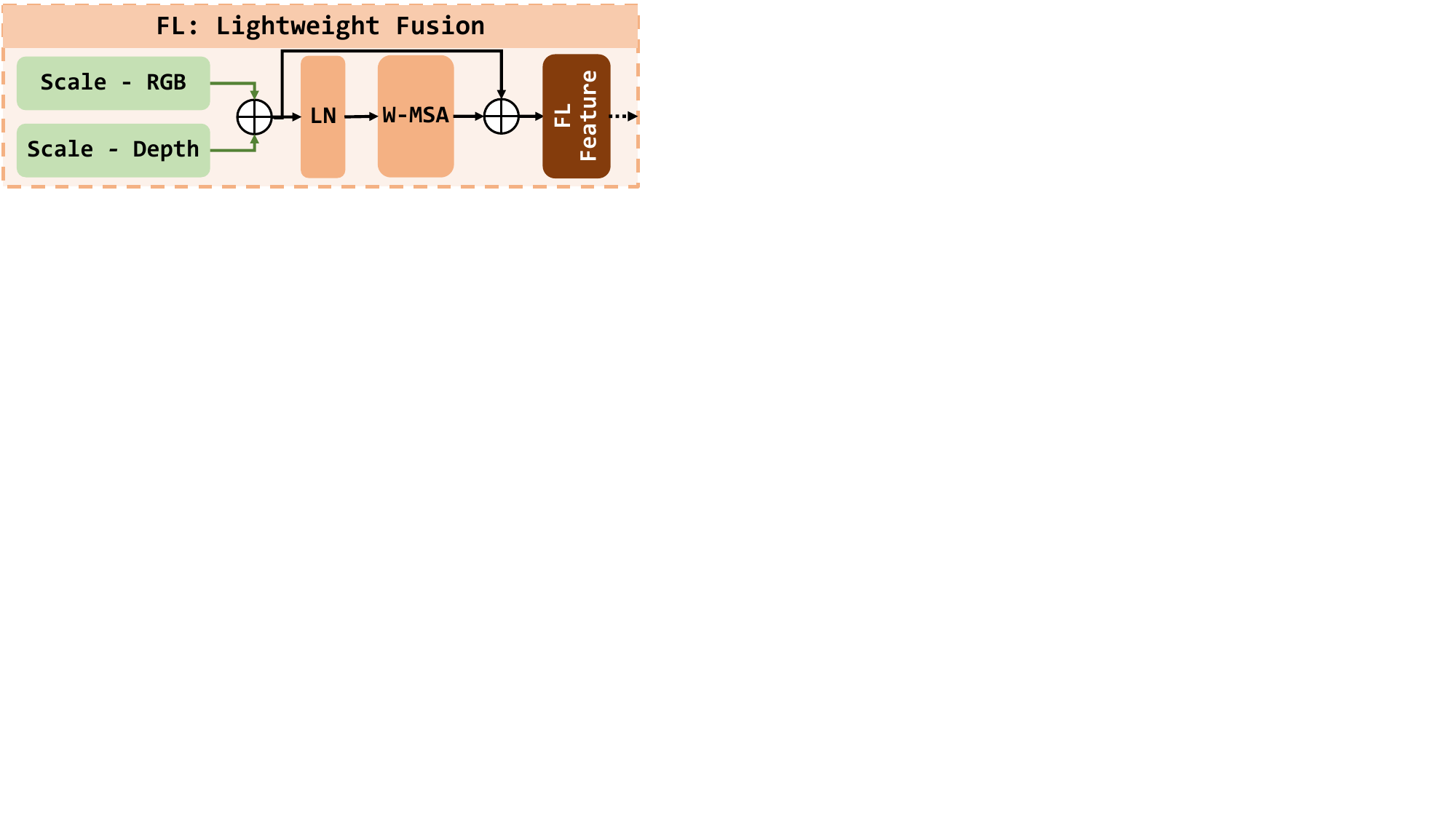}
\caption{An illustration of FL for lightweight feature fusion. FL performs addition operation of the RGB and depth features from the same encoder scale to generate $\mathbf{f}$. The final FL output is the addition of $\mathbf{f}$ and $\mathbf{f}'$, which is generated by an attention module (W-MSA) based on $\mathbf{f}$.}
\label{fig:FL}
\end{figure}

\subsubsection{FL: Lightweight Fusion}
The encoder of NuNet has four scales and each of them generates abstract representations of the input images. We extract such representations from these scales as features $\mathbf{f}_1^\text{R},\ldots,\mathbf{f}_4^\text{R}$ for RGB and $\mathbf{f}_1^\text{D},\ldots,\mathbf{f}_4^\text{D}$ for depth inputs, respectively, where the subscripts indicate the scale identifier. For scale $s$, we perform FL for $\mathbf{f}_s^\text{R}$ and $\mathbf{f}_s^\text{D}$ and generate $\mathbf{f}_s^\text{F}$. The detailed implementation of FL is as follows.
\begin{equation}\label{eq-FL}
    \mathbf{f}_s^\text{F} = \mathbf{f}_s^\text{R} + \mathbf{f}_s^\text{D} + \textsc{W-MSA}\big(\textsc{LN}(\mathbf{f}_s^\text{R} + \mathbf{f}_s^\text{D})\big),
\end{equation}
where LN is the abbreviation of the popular LayerNorm. The implementation includes two parts. One focuses on the raw information of $\mathbf{f}_s^\text{R}$ and $\mathbf{f}_s^\text{D}$ and performs addition operation $\mathbf{f}_s^\text{R} + \mathbf{f}_s^\text{D}$. The other further processes the features based on an attention mechanism, assuming that the attention output is correlated with the nutrition estimation and potentially more related compared to the raw features. While two attention mechanisms are available in our encoder, we propose to use one of them (W-MSA) only to manage the computational overhead. We apply LN to the addition of $\mathbf{f}_s^\text{R}$ and $\mathbf{f}_s^\text{D}$, and use the resulting feature as the input of W-MSA. Overall, FL adds two parts, $\mathbf{f}_s^\text{R} + \mathbf{f}_s^\text{D}$ and the output features from $\textsc{W-MSA}$, as the fused feature $\mathbf{f}_s^\text{F}$.

\subsubsection{FE: Enhanced Fusion}
FL is insufficient to explore the full potential of certain features and we propose FE for enhanced feature fusion in this part. Naturally, FE demands more computing resources compared to FL and excessive usage of it will make a model cumbersome. So, we propose to apply FE for the last encoder scale only with features $\mathbf{f}_4^\text{R}$ and $\mathbf{f}_4^\text{D}$ as shown in Fig. \ref{fig:model}, where such FE block is considered as scale 5. Different from FL's sequential operation flow, FE consists of three parallel operation paths, each processes the features in a unique way. The paths diversify the fusion process and produce intermediate features $\mathbf{f}_5^{\text{F}_\text{C}}$, $\mathbf{f}_5^{\text{F}_\text{M}}$, and $\mathbf{f}_5^{\text{F}_\text{A}}$ based on concatenation, multiplication, and addition operations, respectively. We utilize SwinT's residual connection which is applied to the output of each attention before LN for all the paths. We omit the detailed description of such a connection in the following parts for presentation clarity. With the features from the three paths, we integrate them at the end of the scale and process them with a multi-layer perceptron (MLP) to generate $\mathbf{f}_5^{\text{F}}$. We introduce the technical details of the paths one by one.

\begin{figure}[t]
\centering
\includegraphics[width=0.48\textwidth, trim={0 20em 54em 0}, clip]{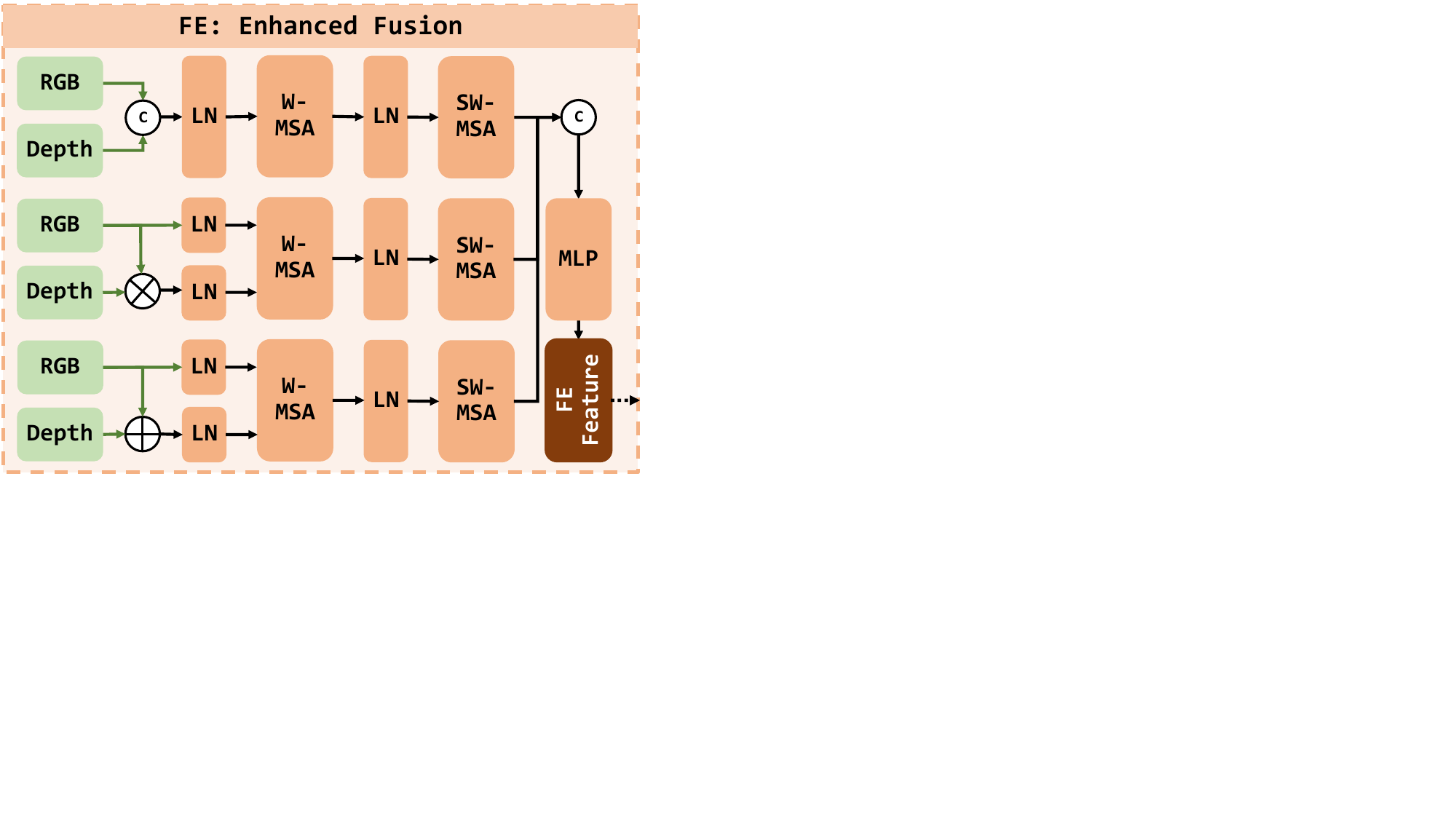}
\caption{An illustration of FE for enhanced feature fusion. FE has three fusion paths for concatenation, multiplication, and addition. Each path utilizes both RGB and depth features in different ways and introduce two attentions (W-MSA and SW-MSA) to process the features. The output of the three paths merge into an MLP before the final FE feature is generated.}
\label{fig:FE}
\end{figure}

\paragraph{Concatenation Path}
The first path generates concatenation-based feature $\mathbf{f}_5^{\text{F}_\text{C}}$. Both $\mathbf{f}_4^\text{R}$ and $\mathbf{f}_4^\text{D}$ are utilized in this path and we propose to process them using two attention mechanisms (W-MSA and SW-MSA) to align with the encoder of NuNet. We expect the usage of two attentions can enhance the feature utilization in FE. Note that the original encoder includes LN and MLP layers before the output of each transformer block and we exclude the layers to facilitate the feature integration which will be detailed later. Here, we represent the operations of this path as follows.
\begin{equation}\label{eq-FE-c}
    \mathbf{f}_5^{\text{F}_\text{C}} = \textsc{SW-MSA}\bigg(\textsc{LN}\Big(\textsc{W-MSA}\big(\textsc{LN}(\mathbf{f}_4^\text{R} \copyright \mathbf{f}_4^\text{D})\big)\Big)\bigg),
\end{equation}
where the input of W-MSA is the concatenation of $\mathbf{f}_4^\text{R}$ and $\mathbf{f}_4^\text{D}$ followed by LN. LN is also applied to the output of W-MSA to produce the input of SW-MSA. Finally, SW-MSA's output feature is considered as $\mathbf{f}_5^{\text{F}_\text{C}}$.

\paragraph{Multiplication Path}
The second path is multiplication-based and generates $\mathbf{f}_5^{\text{F}_\text{M}}$. Different from the concatenation path based on the default W-MSA, this path further modifies the mechanisms for cross-attention between RGB and depth features. To facilitate the description of the cross attention, let us introduce the default attention mechanism first in Eq. (\ref{eq-qkv}).
\begin{equation}\label{eq-qkv}
\textsc{Attention}(Q, K, V) = \mathrm{softmax}(QK^{T} / \sqrt{d} + B)V,
\end{equation}
where $Q$, $K$, and $V\in\mathbb{R}^{n_p, d}$ are the matrices for query, key, and value. Each window has $n_p$ patches, $B$ is a bias component for the window-based attention, and $d$ is the dimension of the query and key. Such attention mechanism commonly takes in features, e.g., the concatenated $\mathbf{f}_4^\text{R}$ and $\mathbf{f}_4^\text{D}$ with LN, to calculate $QKV$ and our attention modification is mainly in such calculation part.

In this paper, we have one fundamental assumption that both RGB and depth features are important whereas RGB plays a dominant role. Nutrition information is considered to be available in RGB images, from which $K$ and $V$ are generated. Generally, $V$ captures the actual nutrition values associated with each part of an RGB image and $K$ is attached to each value as its indicators or keys. The main role of query $Q$ is to find the nutrition-related values based on their keys, where the importance of the values is not necessarily the same. Typically, the query is RGB-based given that both values and keys are from the same RGB source. 

Here, we propose to modify and enhance the query with additional depth information, expecting improved searching efficiency and data understanding. Indeed, this would be intuitively constructive. For example, given an image of food on a plate, the food is closer to a camera compared to the plate surface and the food area can be easily identified with depth. Multiplication serves such a purpose well, utilizing depth to signify spatial differences besides visual variations and highlighting areas where both features are strong. Specifically, we introduce W-MSA$^\times$ and produce $\mathbf{f}_5^{\text{F}_\text{M}}$ as,
\begin{equation}\label{eq-FE-m}
    \textsc{SW-MSA}\bigg(\textsc{LN}\Big(\textsc{W-MSA}^\times\big(\textsc{LN}(\mathbf{f}_4^\text{R} \times \mathbf{f}_4^\text{D}), \textsc{LN}(\mathbf{f}_4^\text{R})\big)\Big)\bigg),
\end{equation}
where there are two input items for W-MSA$^\times$ each with an independent LN. One item is the RGB-only $\mathbf{f}_4^\text{R}$ and the other is the element-wise multiplication of $\mathbf{f}_4^\text{R}$ and $\mathbf{f}_4^\text{D}$, configured with the same length. The new attention mechanism of W-MSA$^\times$ based on the input items is,
\begin{equation}\label{eq-qkv-mul}
\textsc{Attention}(Q^{\text{R}\times \text{D}}, K^\text{R}, V^\text{R}),
\end{equation}
where the modified query is derived from the first input item of W-MSA$^\times$ for both RGB and depth features, and the key $K^\text{R}$ and value $V^\text{R}$ are derived from the second item. Finally, we follow the same as the first path to perform LN for the output of attention W-MSA$^\times$ to produce the input of the next attention SW-MSA.

\paragraph{Addition Path}
The third path is addition-based. It shares a very similar idea as the second path and the main difference is that the multiplication is replaced with an addition operation. Unlike the multiplication path which is nonlinear and highlights the interaction between RGB and depth features, this path is linear and maintains a relatively clear separation between the features with a straightforward aggregation. Specifically, we propose another modified attention W-MSA$^+$ based on which we calculate feature $\mathbf{f}_5^{\text{F}_\text{A}}$ as,
\begin{equation}\label{eq-FE-a}
    \textsc{SW-MSA}\bigg(\textsc{LN}\Big(\textsc{W-MSA}^+\big(\textsc{LN}(\mathbf{f}_4^\text{R} + \mathbf{f}_4^\text{D}), \textsc{LN}(\mathbf{f}_4^\text{R})\big)\Big)\bigg),
\end{equation}
where the first attention W-MSA$^+$ takes in two input items, one is the RGB-only $\mathbf{f}_4^\text{R}$ and the other is the element-wise addition of $\mathbf{f}_4^\text{R}$ and $\mathbf{f}_4^\text{D}$, each item with an LN. The attention then can be represented as,
\begin{equation}\label{eq-qkv-add}
\textsc{Attention}(Q^{\text{R}+ \text{D}}, K^\text{R}, V^\text{R}),
\end{equation}
where the only difference from the attention in Eq. \ref{eq-qkv-mul} is the query, which is based on the addition of RGB and depth features. The output of this attention is used as the input of the second attention SW-MSA after LN and finally the fused feature $\mathbf{f}_5^{\text{F}_\text{A}}$ can be derived.

\paragraph{Fusion of the Paths}
FE generates intermediate features $\mathbf{f}_5^{\text{F}_\text{C}}$, $\mathbf{f}_5^{\text{F}_\text{M}}$, and $\mathbf{f}_5^{\text{F}_\text{A}}$, from its concatenation, multiplication, and addition paths, respectively. We direct them into an MLP for another level of feature fusion and map these features into the fused feature $\mathbf{f}_5^{\text{F}}$ as the output of FE. Worth mentioning that having different paths with different operations is important for NuNet. They together enable rich feature representations to support enhanced image understanding from comprehensive aspects. This brings other benefits such as the model's flexibility and robustness. In our experimental study, we show that our feature fusion modules are effective at utilizing the features derived from both RGB and depth images.

\subsection{Multi-Scale Decoder}
NuNet employs a multi-scale decoder to align with the multi-scale encoder. Such a multi-scale concept can be referred to as deep supervision \cite{xie2015holistically}, which enhances a single-scale decoder with semantically meaningful features in multiple scales. Seen from Fig. \ref{fig:model}, the decoder starts from the output feature of FE in scale 5, i.e., $\mathbf{f}_5^{\text{F}}$. The feature goes into five linear heads, each for one of the five nutritional factors, to generate nutrition estimations. Compared to the beginning scales, FE's $\mathbf{f}_5^{\text{F}}$ is closer to the NuNet output and intuitively better correlated with the final nutrition output of NuNet. The estimation based on $\mathbf{f}_5^{\text{F}}$ is thus considered the most dominant part of the final nutrition estimation of NuNet.

Yet, $\mathbf{f}_5^{\text{F}}$ alone is insufficient to reveal the complex relationships between RGB and depth features in different encoder scales. So we introduce scales $s_6,\ldots,s_9$ in the decoder, each comprising a decoder block based on U-Net \cite{ronneberger2015u} followed by five linear heads. U-Net is chosen because of its effectiveness in context capture and precise localization with its contracting and symmetric expanding paths and special skip connections. However, NuNet is not restricted to U-Net.

Scale 6 is the first of these scales and we let FE's $\mathbf{f}_5^{\text{F}}$ and FL's $\mathbf{f}_4^{\text{F}}$ be concatenated as the input of the scale to well explore the semantic information from the FE module and the FL module for the last encoder scale for both RGB and depth. The U-Net of the scale processes the input and produces $\mathbf{f}_6^{\text{F}}$ with standard average pooling and flattening operations. Then the feature $\mathbf{f}_6^{\text{F}}$ is concatenated with $\mathbf{f}_3^{\text{F}}$ from the second last encoder scale, as the input of the next decoder scale. We apply the same structure for the rest scales and we generalize the input of each decoder scale $s$ for $6\leq s\leq 9$ as $\mathbf{f}_{10-s}^{\text{F}}\copyright \mathbf{f}_{s-1}^{\text{F}}$, e.g., $\mathbf{f}_1^{\text{F}}\copyright \mathbf{f}_8^{\text{F}}$ for the decoder at scale 9. For each scale $s$, we also introduce five linear heads to process feature $\mathbf{f}_{s}^{\text{F}}$. These scales enhance the nutrition estimation based on FE and the output values of all the decoder scales are added together with the FE-based estimates to produce the final nutrition estimation. We specifically design a loss function based on such final estimation to drive the optimization of NuNet with multi-scale features and the function is detailed below.

\subsection{Loss Function} 
We design and implement the loss function of NuNet as,
\begin{equation}
    \ell = \sum_{j=1}^{k}\Big(\frac{1}{m} \sum_{i=1}^{m} \frac{|\hat{y}_{i}^{j} - y_{i}^{j}|}{y_{i}^{j} + 1}\Big),
\end{equation}
where $m$ is the batch size. This loss function is a variant of the mean absolute percentage error (MAPE). For each batch with $m$ images, the loss is calculated as the average percentage error in a modified version, where the ground-truth value is increased by a positive constant 1 in the denominator. The increase ensures a legal fraction with any non-negative nutrition volume. Then, we sum up the loss of $k$ nutrition factors as the total loss $\ell$. Note that we choose percentage error instead of absolute error. This can be justified by the fact that the nutrition factors (e.g., calories and protein) are of different magnitudes \cite{shao2023vision} and the percentage error offers non-biased emphases on different factors.

\begin{table*}[t]
\caption{Comparison study of NuNet and other nutrition estimation models, including four models from existing references and five implemented by us using transfer learning. For each model, we indicate whether RGB or depth features is used with $\checkmark$ and $\times$ indicating true or false, respectively. The estimation results of five nutrients and the average are reported for each comparison model. The best results for each nutrient or mean performance are in highlighted in bold font. NuNet achieves the best mean performance and we report NuNet's improvement to each comparison model in percentage in the last column. \label{tab-comparison}}
\centering
\renewcommand{\arraystretch}{1.3}
\begin{tabular}{c|c|cc|ccccc|cc}
\hline\hline
& ML Backbone & RGB & Depth & Calorie  & Mass  & Fat  & Carbs  & Protein  & Mean & Improvement \\
\hline\hline

\multirow{4}{*}{\parbox{5em}{\centering Results from References}} & Inception \cite{thames2021nutrition5k} & $\checkmark$ & $\times$ & 26.10 & 18.80 & 34.20 & 31.90 & 29.50 & 28.10 & 79.5
\\
& Inception \cite{thames2021nutrition5k} & $\checkmark$& $\checkmark$& 18.80&18.90& \textbf{18.10}&23.80&20.90&20.10&28.4\\
& SwinT \cite{shao2022rapid} & $\checkmark$ &$\times$& 15.30& 12.50& 22.10& 20.80& \textbf{15.40}& 17.20&9.9\\
& ResNet \cite{shao2023vision} & $\checkmark$ &$\checkmark$ & 15.00& 10.80& 23.50& 22.40& 21.00& 18.50&18.2\\
%& \cite{vinod2022image} & $\checkmark$ &$\times$& 13.57& $-$ & $-$ & $-$ & $-$ & $-$  & \\\hline
\hline
\multirow{5}{*}{\parbox{5em}{\centering Transfer Learning}} & ResNet & $\checkmark$ & $\times$ &18.49& 14.36& 26.70& 25.90& 25.71& 22.23&42.0\\
& Inception & $\checkmark$ &$\times$& 18.68& 14.91& 26.21& 26.47& 24.92& 22.24&42.0\\
& VGG & $\checkmark$ & $\times$& 16.80& 12.67& 24.38& 23.57& 23.72& 20.23&29.2\\
& EfficientNet & $\checkmark$&$\times$& 17.28& 14.21& 24.61& 24.71& 23.11&20.78&32.8\\
& SwinT & $\checkmark$&$\times$& 14.39& 11.03& 21.72& 19.93& 20.15& 17.44&11.4\\ \hline
Ours & NuNet & $\checkmark$ & $\checkmark$ & \textbf{12.80}& \textbf{8.72}& 19.67& \textbf{18.66}& 18.42& \textbf{15.65}& $-$\\
\hline\hline
\end{tabular}
\end{table*}

\section{Experimental Study}
\label{sec:exp}
In this section, we present our experiment setup first followed by the experiment results and discussions. 

\subsection{Experimental Setup}
The setup includes three aspects detailed below.

\subsubsection{Dataset and Pre-Processing}
We use the popular \texttt{Nutrition5k} dataset \cite{thames2021nutrition5k} for our experimental study. The dataset has $\sim$3.5 thousand food samples, each of which has both RGB and depth information and provides the ground-truth of the five nutritional factors considered in this paper. We perform data pre-processing to generate a clean dataset for our study. We use the training and testing datasets specified in \cite{shao2023vision}. A minor difference is that the testing dataset from \cite{shao2023vision} has three fewer images, e.g., image \texttt{dish\_1562947503}, which are duplicated in the training datasets. We removed such images in our experiments for comparison with existing works for nutrition estimation.

We perform data augmentation to improve the dataset diversity. We  generate new RGB and depth images by resizing the original ones to 238$\times$238 resolution and then perform a center crop to 224$\times$224 resolution to enlarge the portion of food in the images. We further augment the dataset by flipping the images either vertically or horizontally randomly. Our data augmentation specialized for RGB images includes sharpness adjustment for 0.5\% images by a factor of 2 in each epoch and auto-contrast adjustment for 10\% images. All RGB and depth images are normalized with the default mean and standard deviation in PyTorch.

\subsubsection{Evaluation Metrics}
We aim to approximate the nutrition ground-truth accurately with our proposed NuNet and we adopt two mainstream evaluation metrics in this research domain. The first one is the mean absolute error (MAE) as,
\begin{equation}
\label{eq-mae}
\textsc{MAE}_{j} = \frac{1}{n} \sum_{i=1}^{n} |\hat{y}_{i}^{j} - y_{i}^{j}|,
\end{equation}
where $n$ is the number of testing images for nutritional factor $j\leq k$, e.g., calorie. Another metric is proposed by \cite{thames2021nutrition5k} as a variant of MAPE between $\textsc{MAE}^{j}$ and the mean ground-truth of nutrition factor $j$. We maintain the same metric abbreviation for simplicity and the metric is calculated as,
\begin{equation}
\textsc{MAPE}_{j} = 100 \times \frac{\textsc{MAE}_{j}}{\frac{1}{n} \sum_{i=1}^{n} y_{i}^{j}},
\end{equation}
for nutritional factor $j$ and accordingly the average performance of the $k$ nutritional factors is,
\begin{equation}
\textsc{MAPE} = \frac{1}{k} \sum_{j=1}^{k} \textsc{MAPE}_{j}.
\end{equation}

\subsubsection{Implementation Configurations}
We implement NuNet using PyTorch and utilize the existing pre-trained weights on a PyTorch docker container from \texttt{nvcr.io/nvidia} to speed up the training and optimization. All models are trained on NVIDIA Tesla V100 32GB GPU. Our training strategy includes a batch size of 32, 150 epochs with \texttt{adam} optimizer, 1$\mathrm{e}{-}$4 learning rate, 1$\mathrm{e}{-}$5 weight decay, 1$\mathrm{e}{-}$6 epsilon, and 0.99 exponential decay. %The reported experiment results are based on 10 independent runs by default for cross-validation.

\subsection{NuNet Performance and Comparison Study}
We first want to understand the performance of NuNet compared to other solutions for nutrition estimation. We show the results in Table \ref{tab-comparison}. 

\subsubsection{Comparison Solutions}
Recently, a few papers \cite{thames2021nutrition5k,shao2022rapid,shao2023vision} analysed the \texttt{Nutrition5k} dataset and reported the estimation results. The first paper \cite{thames2021nutrition5k} studied the impact of RGB and depth features on nutrition estimation with the CNN-based Inception models \cite{szegedy2016rethinking} and demonstrated the usefulness of incorporating depth in the estimation models. The rest two papers are from the same research group. One paper \cite{shao2022rapid} considered RGB only and used a transformer and the other \cite{shao2023vision} studied the impact of both RGB and depth with the CNN-based ResNet \cite{he2016deep}. These papers represent the state-of-the-art of this research domain.

Nutrition estimation is image-based and CNN-based models have been widely used in the past decade for image processing. We thus customize several CNN models for nutrition estimation by retraining the models using the nutrition dataset. This is a typical transfer learning approach. Specifically, we investigate four CNN models including ResNet, Inception, VGG \cite{simonyan2014very}, and EfficientNet \cite{tan2019efficientnet}. We further follow the same idea for the transformer-based SwinT. We replace these models' final linear heads with a 4096-dimensional MLP layer. None of these models support multi-modality data (e.g., RGB and depth) by default, and we report the RGB-only performance, which is generally better than depth-only based on our tests.

\subsubsection{Performance Analysis}
Seen from Table \ref{tab-comparison}, a clear observation is that NuNet is highly competitive compared to the comparison models. For mean performance, NuNet is the only one achieving below 16\% percentage error, i.e., 15.65\%. The best of the rest is the model from \cite{shao2022rapid} with 17.20\%, which is 10\% worse than NuNet. For the rest models, the error rates are even higher and NuNet's improvement to them becomes more significant. Such competitive performance of NuNet remains true for each of the five nutrients. Among them, NuNet performs the best for three nutrients including calorie, mass, and carbs. For the rest two, fat and protein, NuNet is the second best. Specifically, NuNet and the Inception model from \cite{thames2021nutrition5k} are the only models with below 20\% error rate for fat estimation. For protein, only NuNet and SwinT from \cite{shao2022rapid} can achieve such a level of performance. 

Another observation is that using both RGB and depth features is beneficial. A direct comparison is between two Inception-based models from \cite{thames2021nutrition5k}, one with RGB only and the other has both. Compared to the former, the latter improves the performance for four out of five nutrients significantly, e.g., 1.9x better for fat. The only exception is mass, where the difference between the two models is merely 0.1\%. The advantage is reflected in the mean error rate, where the latter is 40\% better. We may also see from the comparison between NuNet with both RGB and depth features and the transfer learning-based SwinT with only RGB. The backbone of the two models is the same, yet NuNet outperforms its counterpart for all the five nutrients with an 11\% average error rate improvement. These results demonstrate the effectiveness of using both RGB and depth features for nutrition estimation. 

The results also show the transformer's competitiveness as a backbone. Among five models in the transfer learning category, four of them are CNN-based and SwinT is transformer-based. We can see that SwinT has the best performance in all five nutrients compared to the rest models. The model is the only one in this category achieving below 20\% error rate. This argument can further be justified by NuNet, which is also transformer-based and performs even better. Besides, it is interesting to find out that transformer-based SwinT from the existing literature is more accurate than a ResNet-based model, despite the latter having the input of both RGB and depth features and the SwinT model having RGB only. This further highlights the modelling capability of the transformer for our image-based nutrition estimation.

\subsection{Feature Fusion Analysis}
Feature fusion specialized for nutrition estimation is one of the main contributions of this paper. In this part, we investigate the impact of both FL and FE on the overall estimation performance, and let us start from FL.

\subsubsection{FL Impact Analysis}
To analyze the FL's impact on nutrition estimation performance, we replace it with different fusion modules and we introduce the details of them as below.

\paragraph{Substitute Fusion Modules}
In Eq. \ref{eq-FL}, FL has two parts, one is the aggregated feature by adding RGB and depth features and the other further applies attention to the aggregated feature. A natural comparison is between FL and either the first part or the second part. We report them in the first category of substitutes in Table \ref{tab:FL}. For the second category, we consider increasing the complexity of FL's attention part, by further adding SW-MSA on top of W-MSA or applying a full transformer block with both attention and MLP. The attention can be either addition-based, the same as FL, or multiplication-based. Furthermore, in the next category, we test existing fusion modules sharing minimal similarity with FL. The modules include CDC \cite{li2020icnet}, CBAM \cite{woo2018cbam}, and CRM \cite{ji2021calibrated}, all of which have both addition and multiplication operations for RGB and depth features. The last two correspond to one convolutional layer, based on either the addition or multiplication of the features. Finally, Table \ref{tab:FL} shows the results of NuNet with FL.

\begin{table}[t]
\caption{The performance (MAPE) impact of FL and its substitute fusion modules in NuNet for nutrition estimation. Each module is based on either addition, or multiplication, or both, and we show the number of attentions used in each of them. The size of the modules is measured by the percentage of the module's parameters over all parameters in NuNet. FL contributes to the accurate estimation of NuNet with a moderate size.}
\label{tab:FL}
\centering
\renewcommand{\arraystretch}{1.3}
\def \tmpw{2em}
\begin{tabular}{c|ccc|c}
\hline\hline
\multirow{2}{*}{\parbox{8em}{\centering Fusion Module (Scale 1-4)}} & \multirow{2}{*}{\parbox{3em}{\centering $+/\times$}} & \multirow{2}{*}{\parbox{3em}{\centering Attention}} & \multirow{2}{*}{\parbox{3em}{\centering Para. (\%)}} & \multirow{2}{*}{\parbox{4em}{\centering MAPE (\%)}} \\ 
& & & & \\\hline\hline

Addition & $+$ & 0 & 0.0 & 15.69 \\ 
W-MSA & $\times$ & 1 & 2.1 & 15.74 \\
Multiplication & $\times$ & 0 & 0.0 & 15.74 \\\hline

SW-, W-MSA & $\times$ & 2 & 4.1 & 15.74 \\
SW-, W-MSA & $+$ & 2 & 4.1 & 15.66 \\ 
W-MSA \& MLP & $\times$ & 1 & 6.6 & 15.67 \\ 
W-MSA \& MLP & $+$ & 1 & 6.6 & 15.66 \\ 
\hline 

CDC & $+, \times$ & 0 & 0.5 & 15.67 \\
CBAM & $+, \times$ & 0 & 4.7 & 15.77 \\
CRM & $+, \times$ & 0 & 2.1 & 15.72 \\
Conv & $\times$ & 0 & 0.5 & 15.74 \\
Conv & $+$ & 0 & 0.5 & 15.67
\\ \hline

FL (ours) & $+$ & 1 & 2.1 & \textbf{15.65} \\
\hline\hline
\end{tabular}
\end{table}

\paragraph{Performance Analysis}
Seen from the table, we have the following observations. First, FL is the best. It achieves the lowest MAPE of 15.65\%. When only one part of FL is used, the error increases, e.g., 15.74\%\footnote{Note that the results difference in Tables \ref{tab:FL} and \ref{tab:FE} is less significant than Table \ref{tab-comparison} as only one module of NuNuet is changed in the FL and FE analysis.} for its second part. Replacing addition with multiplication for its first part does not lead to better performance either. With more complex attention, the expectation of an improvement in performance could be reasonable; however, FL outperforms its counterparts with complex attention. One reason is that complex attention involves a significant amount of parameters which require much data and advanced training strategies to reach optimal performance. Real-world applications like nutrition estimation often have various constraints like the availability of sufficient data, as a result, the best performance may not be achieved by the most complex models. Compared to existing modules like CDC and CBAM, FL is more effective, largely because of its compatibility with NuNet, sharing the same backbone. 

We also notice that NuNet's overall performance is correlated with different factors, and neither a too simplified nor a cumbersome module can achieve the optimal performance. FL accounts for about 2\% of the model size of NuNet and this is a moderate percentage. Reducing the percentage below 2\% implies poor feature fusion capability. Big modules can approximate FL's performance but at the cost of much more module parameters, e.g., nearly double for using two attentions. Also, addition is shown to be more effective in FL compared to multiplication. This implies that RGB and depth are complementary to each other and addition helps preserve both information in FL, while multiplication may lead to either excessive or insufficient attention in certain windows.  

\subsubsection{FE Impact Analysis}
FE is expected to play a more important role than FL in feature fusion with more comprehensive and complex fusion operations. In this part, we discuss FE's impact on the overall performance and compare FE with other fusion modules. 

\paragraph{Substitute Fusion Modules}
We have three categories of substitute modules. First, we consider the case where FE is removed from NuNet, and the RGB and depth features from scale 4 are simply concatenated. Then, refer to Fig. \ref{fig:FE}, we modify FE in different ways, including removing the SW-MSA or both W- and SW-MSA for each path and restoring to the original MSA without cross-attention. In the second category, we consider using one of the three paths, i.e., concatenation path, multiplication path, or addition path only. Lastly, we replace the whole module of our FE with existing feature fusion modules including the cross-modality transformer (CMT) from visual saliency transformer \cite{liu2021visual} and the information conversion module (ICM) from ICNet \cite{li2020icnet}. We report the results in Table \ref{tab:FE}.

\begin{table}
\caption{The performance (MAPE) impact of FE and its substitute fusion modules in NuNet for nutrition estimation. Each module involves one or multiple addition, multiplication, and concatenation operations. We show the number of cross attentions in each module, and the module size is measured by the number of the module's parameters. FE contributes to the accurate estimation of NuNet with a manageable size.} 
\label{tab:FE}
\centering
\renewcommand{\arraystretch}{1.3}
\def \tmpw{2em}
\begin{tabular}{c|ccc|c}
\hline\hline
\multirow{2}{*}{\parbox{7em}{\centering Fusion Module (Scale 5)}} & \multirow{2}{*}{\parbox{4em}{\centering $+/\times/\copyright$}} & \multirow{2}{*}{\parbox{4em}{\centering Cross Attention}} & \multirow{2}{*}{\parbox{4em}{\centering \# Para. $(\times 10^6)$}} & \multirow{2}{*}{\parbox{4em}{\centering MAPE $(\%)$}} \\
&&&&\\\hline \hline

$\mathbf{f}_4^\text{R} \copyright \mathbf{f}_4^\text{D}$ & $\copyright$ & 0&203.8& 15.85 \\
w/t SW-MSA & $+,\times,\copyright$ & 2& 235.3& 15.70 \\
w/t W-, SW-MSA & $+,\times,\copyright$ & 0& 210.1& 15.81 \\
Original MSA & $+,\times,\copyright$ & 0& 260.5& 15.84 \\ \hline

Concatenation Path & $\copyright$ & 0&239.5& 15.79 \\
Multiplication Path & $\times$ & 1&212.2& 16.10 \\
Addition Path & $+$ & 1&212.2& 16.15 \\ \hline

CMT & $-$ & 2 & 371.8& 15.90 \\
CMT \& MLP & $-$ & 2 & 373.9 & 16.14 \\
ICM & $\copyright$ & 0 & 210.7& 15.84 \\
ICM \& MLP & $\copyright$ & 0 & 214.9 & 16.03\\ \hline

FE (Ours) & $+,\times,\copyright$ & 2& 260.5& \textbf{15.65} \\\hline
\hline
\end{tabular}
\end{table}

\paragraph{Performance Analysis}
As seen from the table, our overall conclusion is that FE performs the best. Compared to the basic concatenation, FE with attention mechanisms analyzes the aggregated features better with improved accuracy. Removing the second attention (SW-MSA) from each of FE's paths hurts FE's accuracy and the accuracy further declines when the first attention (W-MSA) is also removed. FE's cross-attention mechanism seems especially important as even its variants with reduced usage of MSA outperform the original MSA from SwinT. All FE's three paths are important and using one of them only fails to perform as competitive as FE. Among the paths, the concatenation path alone seems more correlated with nutrition estimation compared to multiplication and addition. This is reasonable as both RGB and depth features are retained without any information loss. When FE has only the multiplication or addition path, the error rate increases to over 16\%. 

Existing modules cannot outperform FE. CMT from the visual saliency transformer shares similarities with FE and our encoder. It includes cross-attention and employs a window-based approach where an image is split into windows. One difference is that windows in CMT are overlapped while FE based on SwinT follows a non-overlapped strategy. This might introduce weakened compatibility with NuNet which follows SwinT's strategy in different NuNet components by default. We investigate if the compatibility limitation can be bypassed, by introducing an MLP after the CMT module for feature transformation. The modification however cannot improve CMT's performance. The results of another module ICM are similar. ICM involves three paths similar to FE but attention is not part of any paths. The paths are convolutional based, where the first path processes the concatenation of RGB and depth, and the rest two treat RGB and depth equally before the convolution operations. The results imply the importance of attention mechanisms and FE's emphasis on RGB-dominance, without which ICM cannot reach the same level of performance as FE, even with additional MLP included in ICM.

\subsection{Multi-Scale Encoder and Decoder Analysis}
Besides feature fusion, another important contribution of NuNet is its multi-scale encoder and decoder. In this part, we investigate the impact of multi-scale on NuNet's nutrition estimation performance. 

\subsubsection{The Most Dominant Scale}
Among multiple scales, we aim to understand the importance of different scales and identify the one with the biggest impact on the performance. In NuNet, nutrition estimation as the final output is the summation of the output of each of the multiple decoder scales (from scale 5 to 9). The first decoder scale is based on the feature from FE only and the rest use the features from both FE and FL, directly or indirectly. We calculate the percentage of each scale's values, which can be extracted after the linear heads, over the final estimation values, and we report the results in Table \ref{tab:dominant-scale}.

\begin{table}[t]
\caption{The importance of each decoding scale in NuNet's final nutrition estimation, indicated by the percentage of each scale's estimation values over the final estimation values. The FE-generated feature dominates the estimation and the following scales calibrate the estimation.}
\label{tab:dominant-scale}
\centering
\renewcommand{\arraystretch}{1.3}
\begin{tabular}{cc|rrrrr}
\hline\hline
\multirow{2}{*}{\parbox{3em}{\centering Scale}} & \multirow{2}{*}{\parbox{3em}{\centering Fusion}} & \multicolumn{5}{c}{Percentage (\%)}\\ \cline{3-7}
& & Calorie & Mass & Fat & Carb & Protein \\ \hline\hline

5 & FE & 97.03 & 93.66 & 98.18 & 94.96 & 97.88 \\
6 & FL \& FE & 1.28 & 0.76 & $-$0.02 & $-$0.01 & $-$0.02 \\
7 & FL \& FE & 5.29 & 10.81 & 0.02 & 0.67 & 0.34 \\
8 & FL \& FE & 0.34 & 0.58 & 8.09 & 0.06 & $-$0.02 \\
9 & FL \& FE & 0.14 & 0.15 & 0.13 & 0.15 & 0.04 \\
6$-$9 & FL \& FE & 2.97 & 6.34 & 1.82 & 5.04 & 2.12 \\\hline
\hline
\end{tabular}
\end{table}

From the table, we can observe that scale 5 with the FE feature dominates NuNet's estimation. It forms over nine-tenths of the final estimation, and the percentage can be as high as 97.9\% for protein. For all the rest scales, the summed percentage is 3.6\% on average and as low as 1.8\% for fat. We interpret that NuNet estimates nutritional factors based on the FE feature mainly and uses the rest scales to calibrate the estimation. This is evidenced by the fact that the percentage can be negative for these scales, implying that they tend to minimize the over- or under-estimation. The magnitude of such calibration is expected to be minimized in later scales, and this largely aligns with the results in the table. One exception is scale 6, with a lower significance than scale 7 in most nutrients. The diminished significance could potentially be attributed to its utilization of features from both scale 4 for FL and scale 5 for FE, leading to redundancy as FE has already incorporated the features from scale 4 in estimation. 

% \begin{table}
% \caption{FE as a single-scale decoder. \label{tab:scale6_deep_supervision}}
% \centering
% \renewcommand{\arraystretch}{1.3}
% \begin{tabular}{c|c|cc}
% \hline\hline
% \multirow{2}{*}{\parbox{5em}{}} & \multirow{2}{*}{\parbox{7em}{\centering Fusion Module (Scale 5)}} & \multicolumn{2}{c}{MAPE (\%)}\\ \cline{3-4}
%  & & Single-Scale  & Multi-Scale \\ \hline\hline

% \multirow{2}{*}{\parbox{5em}{\centering References}} & CMT & 21.94 & 15.90 \\
% & ICM & 20.36 & 15.84 \\\hline

% Ours & FE & 15.71 &
% \textbf{15.65} \\
% \hline\hline
% \end{tabular}
% \end{table}

\subsubsection{Single-Scale vs. Multi-Scale}
\begin{figure}[t]
\centering	
\def \tmpw{0.335}
\subfigure[NuNet (Ours)]{\includegraphics[width=\tmpw\linewidth]{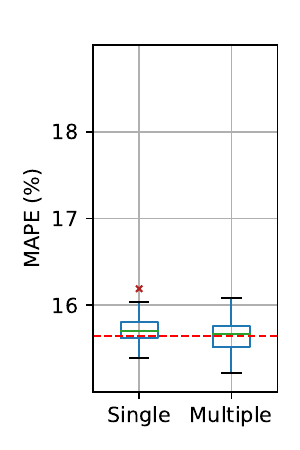}\label{fig:multi-scale-nunet}}
\hspace{-0.75em}
\subfigure[CMT]{\includegraphics[width=\tmpw\linewidth]{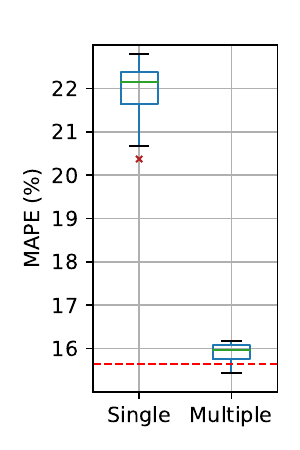}\label{fig:multi-scale-cmt}}
\hspace{-0.75em}
\subfigure[ICM]{\includegraphics[width=\tmpw\linewidth]{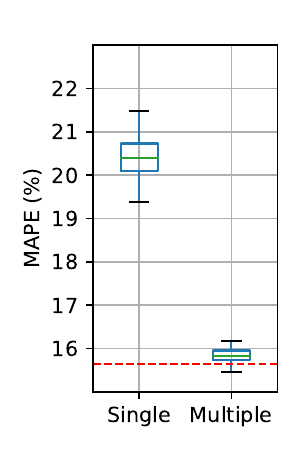}\label{fig:multi-scale-icm}}
\caption{The performance comparison of single-scale and multi-scale based models, where the single-scale implementations involve the FE feature only. FE is also replaced with CMT or ICM to further demonstrate the impact of having multiple scales. Overall, multi-scale implementations outperform their single-scale counterparts significantly. The red horizontal line is a reference line corresponding to NuNet's performance of 15.65\% MAPE.}
\label{fig:multi-scale}
\end{figure}

Besides the importance of each scale, we also investigate the overall impact of multi-scale and we compare it with the single-scale implementation (with scale 5 for FE only). With the removal of scales 6- to 9, the FL features from scales 1 to 4 are excluded automatically. To better evaluate the effect of having a multi-scale, we further perform a comparison by replacing FE with the popular CMT or ICM in NuNet. The statistical results are shown in Fig. \ref{fig:multi-scale}. 

We can see that multi-scale contributes to significantly improved estimation performance. In NuNet, multi-scale helps push the error rate lower with single-scale generally leading to a higher error rate and more significant uncertainty (noticed from the height of the boxes and outlier). CMT and ICM show a similar pattern. For single-scale based on CMT, the estimation MAPE is 21.9\% and multi-scale allows for a big improvement to 15.9\%. For ICM, it has 20.4\% and 15.8\% MAPE for single-scale and multi-scale, respectively. Note that both CMT and ICM show a much larger performance gap between single-scale and multi-scale. This may highlight the importance of multi-scale if a highly effective feature fusion module is not part of the model.

\section{Conclusion}
\label{sec:conclusion}
In this paper, we propose NuNet, which is a transformer-based network for nutrition estimation. NuNet utilizes both RGB and depth information from food images and extracts features with its unique multi-scale architecture and feature fusion modules. Specifically, we develop both FL and FE, with the former specialized for lightweight feature fusion for different encoding scales and the latter customized for comprehensive feature extraction of the last encoder scale. We further match the multiple scales of the encoder with a multi-scale decoder for deep supervision. Our experimental results show that NuNet is highly competitive, achieving an error rate of 15.65\% only for nutrition estimation, and outperforms all its variants and existing solutions considered in this study. We also demonstrate the effectiveness of feature fusion and our multi-scale architecture that contribute to the remarkable performance of NuNet. Overall, NuNet is effective and efficient for nutrition estimation and dietary management. It highlights the importance of using both RGB and depth information as well as developing specialized models that could inspire further applications not limited to dietary management. 

\balance
\bibliographystyle{IEEEtran}%ieeetr IEEEtran IEEEtranS
\bibliography{ref}

\end{document}